\pdfoutput=1

\documentclass[11pt]{article}

\usepackage{xcolor}
\usepackage{graphicx}
\usepackage{booktabs}
\usepackage{comment}
\usepackage{multirow}
\usepackage[]{acl}

\usepackage{times}
\usepackage{latexsym}

\usepackage[T1]{fontenc}

\usepackage[utf8]{inputenc}

\usepackage{microtype}
\usepackage{lineno}

\title{UBERT: A Novel Language Model for Synonymy Prediction at Scale in the UMLS Metathesaurus}

\author{Thilini Wijesiriwardene$^1$\thanks{~~This work was conducted while the author was an intern at the National Library of Medicine} , Vinh Nguyen$^{2}$, Goonmeet Bajaj$^3$, Hong Yung Yip$^1$, Vishesh Javangula$^4$\\
\bf{Yuqing Mao$^2$, Kin Wah Fung$^2$, Srinivasan Parthasarathy$^3$, Amit Sheth$^1$, Olivier Bodenreider$^2$} \\
  $^1$AI Institute, University of South Carolina, 
  $^2$National Library of Medicine\\
  $^3$The Ohio State University, 
  $^4$George Washington University\\
  \texttt{\{thilini, amit\}@sc.edu},  
  \texttt{\{vinh.nguyen, yuqing.mao, kfung\}@nih.gov}, \\
  \texttt{\{bajaj.32,parthasarathy.2\}@osu.edu}, 
  \texttt{hyip@email.sc.edu}, \\
  \texttt{visheshj123@gwu.edu, obodenreider@mail.nih.gov} 
  }

\begin{document}
\nolinenumbers
\maketitle
\begin{abstract}

The UMLS Metathesaurus integrates more than 200 biomedical source vocabularies. During the Metathesaurus construction process, synonymous terms are clustered into concepts by human editors, assisted by lexical similarity algorithms. This process is error-prone and time-consuming. Recently, a deep learning model (LexLM) has been developed for the UMLS Vocabulary Alignment (UVA) task. This work introduces UBERT, a BERT-based language model, pretrained on UMLS terms via a supervised Synonymy Prediction (SP) task replacing the original Next Sentence Prediction (NSP) task. The effectiveness of  UBERT for UMLS Metathesaurus construction process is evaluated using the UMLS Vocabulary Alignment (UVA) task. We show that UBERT  outperforms the LexLM, as well as biomedical BERT-based models. Key to the performance of UBERT are the synonymy prediction task specifically developed for UBERT, the tight alignment of training data to the UVA task, and the similarity of the models used for pretrained UBERT.

\end{abstract}

\section{Introduction}

The Unified Medical Language System (UMLS) Metathesaurus is a large  biomedical thesaurus developed by the US National Library of Medicine\footnote{\url{https://uts.nlm.nih.gov/}}. It clusters synonymous terms from different biomedical source vocabularies into concepts. The current UMLS Metathesaurus construction process relies heavily on lexical similarity algorithms to identify candidates for synonymy and the final decision for synonymy or non-synonymy among terms comes from the domain experts through manual curation. Given the current scale of the UMLS Metathesaurus, with millions of terms from 214 source vocabularies, it is shown that the current construction process is undoubtedly costly and error-prone \cite{cimino1998auditing, cimino2003consistency, jimeno2009reuse, morrey2009neighborhood, mougin2009analyzing}.\\
\textbf{Motivation.} Clustering biomedical terms into concepts in the UMLS Metathesaurus was formalized into a vocabulary alignment problem identified as UMLS Vocabulary Alignment (UVA) or synonymy prediction task by \cite{10.1145/3442381.3450128}. The UVA is different from other biomedical ontology alignment efforts by the Ontology Alignment Evaluation Initiative (OAEI) due to the extremely large problem size of the UVA with the need to compare 8.7M biomedical terms pairwise (as opposed to tens of thousands of pairs in OAEI datasets). The authors of \cite{10.1145/3442381.3450128} also introduced a scalable supervised learning approach based on the Siamese neural architecture which leverages the lexical information present in the terms.\\
Bidirectional Encoder Representations from Transformers (BERT) \cite{devlin2019bert} is a language model (LM), based on the multi-layer, bidirectional architecture of Transformers \cite{vaswani2017attention}. BERT is originally trained on two self-supervised tasks named Masked Language Modelling (MLM) and Next Sentence Prediction (NSP). Recently BERT has been pretrained on several biomedical and clinical corpora resulting in models, such as BioBERT \cite{10.1093/bioinformatics/btz682}, BlueBERT \cite{peng2019transfer}, SapBERT \cite{liu2021self} and UmlsBERT \cite{michalopoulos2020umlsbert}, which have been used successfully on several biomedical NLP tasks, such as  biomedical named entity recognition, biomedical relation extraction, biomedical question answering, biomedical sentence similarity, biomedical document classification and medical entity linking to provide state-of-the-art (SOTA) results. We believe that a parallel can be drawn between Entity Linking (EL) and UVA, because both tasks try to link an entity to a specific term in a reference terminology. The difference is that, in EL, the entity to be linked is found in context (embedded in a sentence or paragraph), whereas, in UVA, the entity is provided without any context (i.e., just the term itself). Our motivation for this work is to investigate how BERT, pretrained on similar data (i.e.,  UMLS data and biomedical literature) performs in the context of UVA.\\
\textbf{Objectives.}
The first objective of this work is to improve upon the performance of current baselines for the UVA task. To this end, we develop UBERT, a novel BERT-based language model specifically trained for synonymy prediction.\\
The second objective is to assess the contribution of two elements of UBERT, namely whether the MLM is beneficial and which datasets provide optimal training for the MLM.\\
The third objective is to explore how UBERT performs when further pretrained on several BERT-based models initially pretrained on a variety of biomedical data (BioBERT), clinical data (BlueBERT) and UMLS data (SapBERT, UmlsBERT).\\
Our last objective is to assess the generalizability of UBERT to the entire UMLS Metathesaurus, by analyzing whether overall performance gains realized by UBERT over baselines are conserved across the entire testing dataset.\\ 
\textbf{Approach.} We identify BERT-based models (in this work BERT-based models refer to BioBERT, BLUEBERT, SapBERT and UmlsBERT) and use them as baselines without further pretraining or fine-tuning on the UVA task. Another baseline used in our work is the LexLM provided by \cite{10.1145/3442381.3450128}. Then we design experiments to pretrain UBERT from scratch (without using any trained weights from other biomedical or clinical BERT-based models) resulting in three variants of UBERT. We evaluate the performance of each variant on a test dataset provided in section \ref{synonymy-dataset}. In addition, we further pretrain UBERT on top of already trained weights from four existing BERT-based models and evaluate their performance on the same test datasets. Finally, we perform a semi-qualitative analysis of the performance of UBERT on the testing dataset by computing the usual performance metrics for specific subsets of the testing dataset across the spectrum of lexical similarity between terms in the pairs of terms evaluated for synonymy.\\ 
\textbf{Contributions.} We introduce UBERT, a novel BERT-based language model, and three variants of UBERT based on the pretraining tasks and pretraining data used. We show that SapBERT+UBERT outperforms the previous LexLM baseline and "off-the-shelf" BERT-based baselines. We also demonstrate that, for the UVA task, without further pretraining with UBERT, "off-the-shelf" BERT-based models perform poorly.\\
We show that pretraining with the MLM task first and then pretraining with the SP task results in better performance compared to UBERT without the MLM task. And we further demonstrate that UBERT performs better when the MLM task is trained with UMLS data only (without biomedical literature data).\\
We demonstrate that UBERT variants that are further pretrained on BERT-based models perform better than the variants that are not. Further, we show that, among the various biomedical BERT-based models used for pretraining, SapBERT yields the best performance.\\ 
We show that overall performance gains (F1 score) realized by UBERT over baselines are conserved across the entire testing dataset across the spectrum of lexical similarity between terms in the pairs of terms evaluated for synonymy, indicating that UBERT performance is likely to generalize to the entire UMLS Metathesaurus.
\section{Background}
Nguyen et al. \cite{10.1145/3442381.3450128} have elaborated the background knowledge required to understand the UVA task. In this section we will  briefly summarize it. In this work, we use the 2020AA version of the UMLS Metathesaurus which contains 15.5 million \textbf{\textit{atoms}}, the building block of the UMLS Metathesaurus, from 214 souce vocabularies grouped into 4.28 million concepts. An atom (atom string) coming from a source vocabulary is uniquely identified in the UMLS Metathesaurus by an atom unique identifier (AUI). The same term can appear in the UMLS Metathesaurus with different AUIs if it comes from different source vocabularies. Atoms that have the same meaning are clustered into the same concept identified by a concept unique identifier (CUI).
In the UVA task, given two atom strings, a computational model is expected to predict their synonymy (or non-synonymy).

The UMLS Metathesaurus contains approximately ten million English atom strings, each of which being linked to a concept. 
Since the authors of \cite{10.1145/3442381.3450128} focus on assessing whether two atoms are synonymous and should be associated with the same concept, the problem is formulated as a similarity task. We maintain this same problem definition from \cite{10.1145/3442381.3450128}.

\section{Related Work}
\vspace{-3mm}
In this section we briefly review previous work on the UVA task, BERT and how BERT-based LMs are used in BioNLP tasks.

\subsection{LexLM for the UVA Task} 
\vspace{-2mm}
Nguyen et al. \cite{10.1145/3442381.3450128} have introduced UVA as a new task in the BioNLP domain and demonstrated that LexLM, a Siamese architecture-based Bidirectional Long Short Term Memory (Bi-LSTM) network with BioWordVec embeddings \cite{zhang2019biowordvec}. LexLM has a F1-score of 94.8\%, precision of 94.64\%, recall of 94.96\% and outperforms a rule-based approach (RBA) described in the same work, in F1 score (+14.1\%), precision (+2.4\%) and recall (+23\%). 
\subsection{BERT: Bidirectional Encoder Representations from Transformers}

BERT \cite{devlin2019bert} is a language model, based on the multi-layer, bidirectional architecture of Transformers  \cite{vaswani2017attention}, which provides contextual word representations as opposed to  context independent distributed word representations introduced by Word2Vec \cite{mikolov2013distributed}, Glove \cite{pennington-etal-2014-glove}, fasttext \cite{bojanowski2017enriching} and Biowordvec \cite{zhang2019biowordvec} (in the biomedical context). BERT is trained on two unsupervised training tasks, namely Masked Language Modeling (MLM) and Next Sentence Prediction (NSP). 

The MLM task allows the model to learn the bidirectional context of a target word in the training process. An input sequence is passed to the model with 15\% of the tokens masked and the masked tokens are predicted by the model. In order to reduce the mismatch between training and testing data, a masked word is replaced by a [MASK] token only 80\% of the time. Ten percent of the time, the masked word is replaced by a random word and the remaining 10\% of the time, the masked word is unchanged.

The NSP task allows the model to learn the relationship between two consecutive segments of a document (e.g., consider segment A and segment B). This is configured as a binarized classification task where 50\% of the time, segment B actually follows segment A in a document and in the other 50\% it does not.

\subsection{BERT-based Language Models and Biomedical NLP (BioNLP) tasks}


In the biomedical domain, BERT is pretrained on large biomedical corpora to create language models (presented below) that have performed successfully on downstream BioNLP tasks, such as named entity recognition, natural language inference and entity linking. This demonstrates the importance of pretraining BERT-based models on domain specific data to achieve better performance.

SapBERT \cite{liu2021self} and UmlsBERT \cite{michalopoulos2020umlsbert} are two recent BERT-based models that leverage UMLS Metathesaurus data for pretraining BERT. SapBERT pretrains on synonymous and non-synonymous pairs of English entries in the UMLS Metathesaurus belonging to the same concept for the downstram task of Medical Entity Linking. The authors have introduced a metric learning framework to self-align the synonymous biomedical entities. UmlsBERT augments the MLM task for pretraining with UMLS Metathesaurus terms by taking into consideration the associations between the words specified in the UMLS Metathesaurus. Instead of predicting a single word in the MLM task, UmlsBERT tries to predict all the acceptable words for the masked token through words associated with the same CUI.





\section{UBERT}
UBERT is the novel BERT-based LM architecture we are introducing. In the subsequent subsection, we describe the novel additions we made to BERT to create UBERT as well as the datasets used to train UBERT.

\subsection{UBERT Architecture}

 \begin{figure}[htp]
     \centering
     \includegraphics[width=8cm]{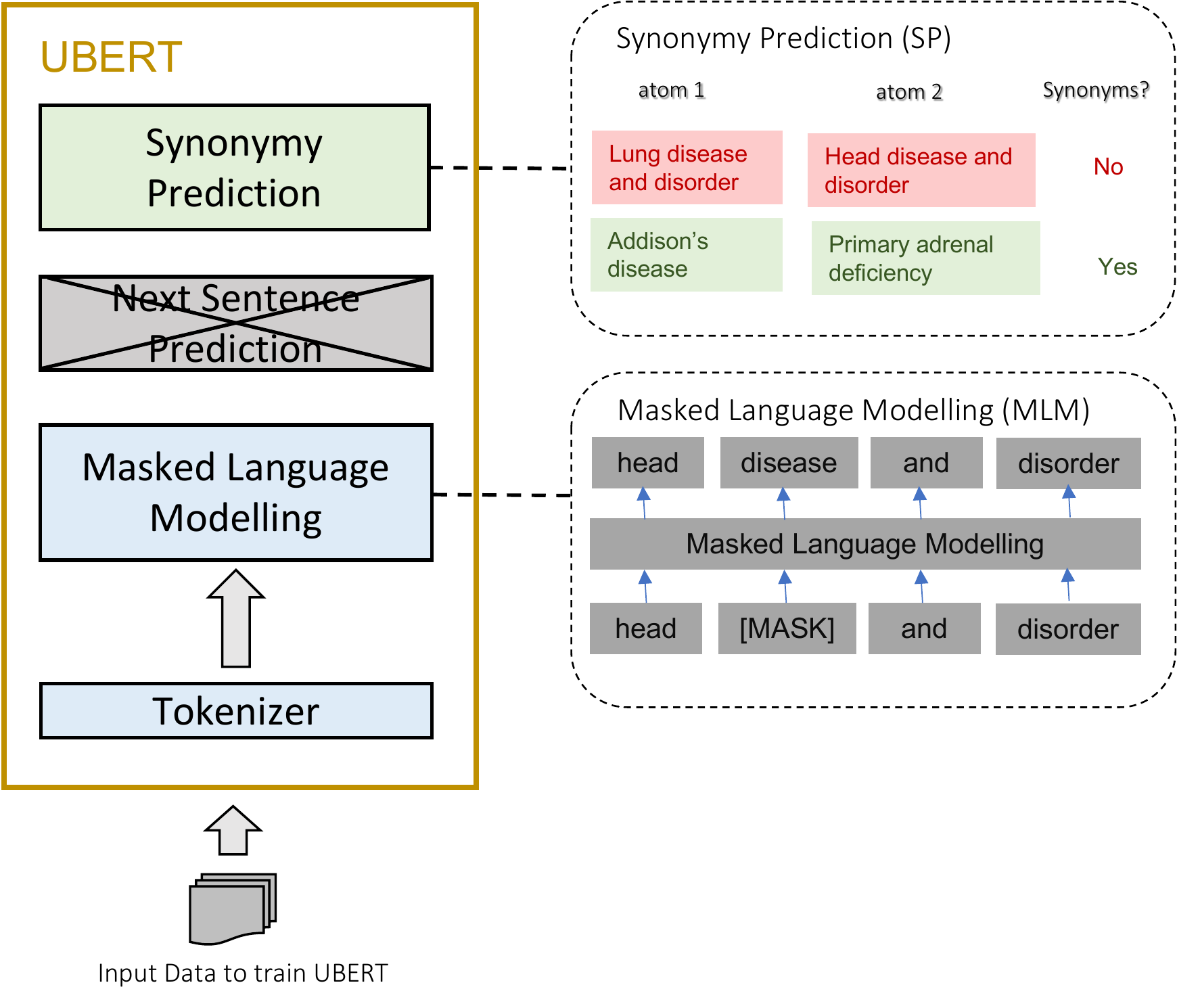}
     \caption{General UBERT Architecture; In UBERT-A, Masked Language Modelling is not used and in UBERT-B1 and UBERT-B2, Masked Language Modelling is used, but trained on different biomedical datasets}
     \label{fig:ubert-archi}
 \end{figure}

As illustrated in Figure \ref{fig:ubert-archi}, we use the MLM task as it is and change the NSP task to a binarized synonymy prediction task when pretraining UBERT. Pretraining on the MLM task is directly inherited from the original BERT architecture\footnote{We have used the Transformers implementation of BERT for pretraining}. Datasets used for pretraining and testing are presented in section \ref{datasets}

\subsubsection{Synonymy Prediction (SP)}
We are re-purposing the binarized classification task of NSP to SP. This is a supervised task whereas the original NSP was an unsupervised task. Atom string pairs annotated as synonymous or non-synonymous are used as training data in the pretraining process. Two atoms are considered synonymous if they belong to the same concept and non-synonymous otherwise. In place of sequence 1 and sequence 2 in NSP, we use atom string 1 and atom string 2 and in place of next sentence label, the state of synonymy (or non-synonymy) between atom string 1 and atom string 2 is used (binary label [0 or 1]). Similar to NSP, where special [CLS] and [SEP] tokens are used to separate two input sequences, in SP we use them to separate the two atom strings.\\
The input is processed as following for UBERT's SP task before it is sent through the model. A [CLS] token is added to the beginning of the first atom string and a [SEP] token is added to the end of each atom string. Another embedding indicating atom string 1 or atom string 2 is then added to each token. Finally a positional embedding is added to tokens indicating the position of each token. This processing is similar to how BERT preprocesses its input and we direct the reader to \cite{devlin2019bert} for a full explanation of the concept and implementation.\\
When predicting whether two atom strings are synonymous or not, the following actions are taken by UBERT. (1) The input sequence presented above is sent through the UBERT model. (2) The output of the [CLS] token is then transformed to a $2 X 1$ vector using a fully connected, binary classification layer. (3) Finally, to calculate the probability of synonymy, the output of the classification layer is sent through a softmax function.\\

\subsubsection{\label{tokenizer}Tokenizer}

Input sequences to both MLM and SP tasks of all the UBERT variants are tokenized using Wordpiece tokenization approach \cite{wu2016google} with a 50000 token vocabulary. The tokenizer  was trained on UMLS atom strings described in section \ref{umls-atom-strings} and a biomedical literature dataset described in section \ref{literature-data}. For other BERT-based models, the tokenizers provided online by the respective authors were used. 

We combine both UMLS atom strings and the biomedical literature when training the tokenizer, because we have identified that 56\% of the words in the UMLS are not found in the biomedical literature and 86\% of the words in the biomedical literature are not found in the UMLS. 

\subsection{\label{datasets}Datasets}
This section discusses the three datasets used in the pretraining and testing of the UBERT variants.

\subsubsection{\label{umls-atom-strings}UMLS atom strings dataset}

This consists of 8,713,194 English UMLS atom strings extracted from the 2020AA release of the UMLS Metathesaurus. 

\subsubsection{\label{literature-data}Biomedical literature dataset}

In this work we use the dataset of PubMed abstracts and PubMed Central (PMC) full-text articles provided by \cite{10.1093/bioinformatics/btz682} with 4.5 billion and 13.5 billion words respectively.

\subsubsection{\label{synonymy-dataset}Annotated synonymy datasets}

We thank \cite{10.1145/3442381.3450128} for providing the training, development and testing datasets used in this work. These datasets consist of English atom strings from active source vocabularies of the 2020AA release of the UMLS Metathesaurus. Annotated datasets are constructed by including synonymous atom string pairs (atom strings linked to the same concept) and non-synonymous atom string pairs (atom strings linked to different concepts). There are approximately 27.9M synonymous pairs (positive samples) and $10^{14}$ non-synonymous atom pairs (negative samples). The ratio between non-synonymous atom string pairs and synonymous atom string pairs is high since most atoms do not share the same CUI. Therefore to create more balanced datasets \cite{10.1145/3442381.3450128} have reduce the negative (non-synonymous) samples to approximately 170M.

In this work, we use the GEN\_ALL dataset from \cite{10.1145/3442381.3450128}. The training and testing datasets do not contain overlapping data points. The training  dataset consists of 118,789,005 annotated (for synonymy and non-synonymy) atom string pairs and testing dataset consists of 171,991,918 annotated atom string pairs. Statistics of the training, development and testing datasets are listed in Table \ref{tab:annotated-synonymy-dataset}.


\begin{table}[ht]
\resizebox{0.5\textwidth}{!}{%
\begin{tabular}{@{}crrr@{}}
\toprule
\multicolumn{1}{l}{\textbf{}} & \multicolumn{1}{c}{Training} & \multicolumn{1}{c}{Development} & \multicolumn{1}{c}{Testing} \\ \midrule
Synonyms & 16,743,627 & 5,581,208 & 5,581,208 \\ \midrule
Non-synonyms & 102,045,378 & 34,015,125 & 166,410,710 \\ \midrule
Total & 118,789,005 & 39,596,333 & 171,991,918 \\ \bottomrule
\end{tabular}%
}
\caption{Number of synonymous and non-synonymous atom string pairs in the training, development and testing dataset (GEN\_ALL).}
\label{tab:annotated-synonymy-dataset}
\end{table}

\vspace{-7mm}

\section{Experimental Setup and Evaluations}

 \begin{figure*}[ht]
     \centering
     \includegraphics[width=16cm]{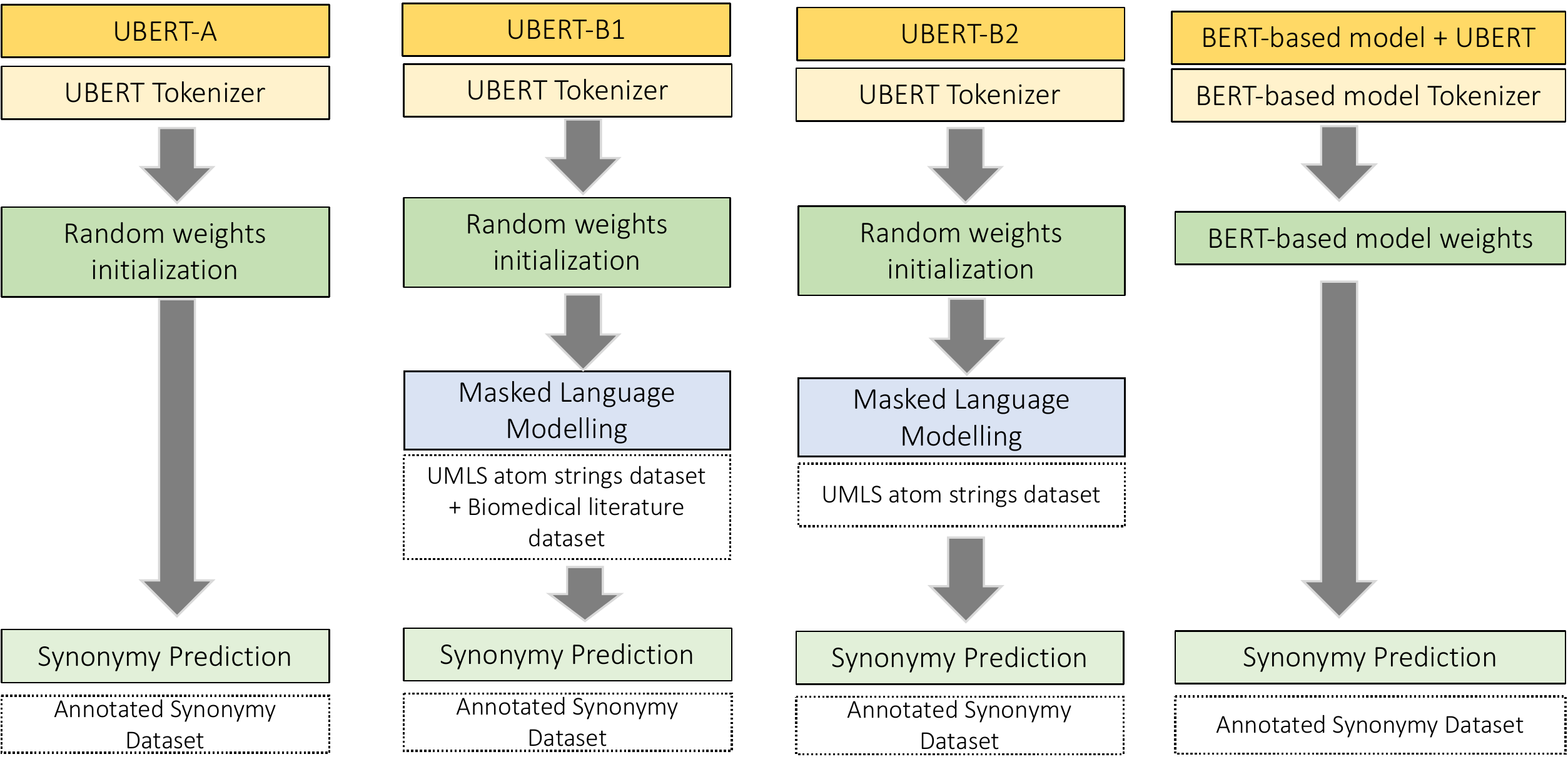}
     \caption{Experimental setup for training UBERT variants and pretrained variants. The datasets used for pretraining are indicated inside the dotted line boxes}
     \label{fig:experimental-setup}
 \end{figure*}
 
 \begin{table*}[ht]
\centering
\resizebox{1\textwidth}{!}{%
\begin{tabular}{@{}cccccc@{}}
\toprule
Model              & GPUs                                             & Batch Size per GPU & Input Sequence Length & Num. of Days Trained & Num. of Epochs/ Steps Trained \\ \midrule
UBERT-A            & 16 v100x GPUs (32GB of  RAM) & 256 & 32 & 8 & 50 epochs  \\ \midrule
UBERT-B1 (MLM task) & \multicolumn{1}{l}{16 v100x GPUs (32GB of  RAM)} & 8                  & 512                   & 7                    & 410 steps                     \\ \midrule
UBERT-B2 (MLM task) & 16 v100x GPUs (32GB of  RAM) & 256 & 32 & 8 & 3.5k steps \\ \bottomrule
\end{tabular}%
}
\caption{Resource utilization}
\label{tab:computation-resources}
\end{table*}

In this section we present the pretraining and evaluation setup of UBERT variants and pretrained variants (see Figure \ref{fig:experimental-setup}). 

\vspace{-1mm}
\subsection{UBERT Variants}
We create three UBERT variants, UBERT-A, UBERT-B1 and UBERT-B2 depicted in Figure \ref{fig:experimental-setup}. UBERT-A only uses the SP task, while the other two variants also use the MLM task for pretraining. The difference between UBERT-B1 and UBERT-B2 lies in the dataset used to pretrain the MLM task.

\subsubsection{UBERT-A}
This variant of UBERT is pretrained using only the SP task, i.e., without the MLM task. 
We first initialize UBERT-A with random weights and further pretrain it with the SP task on annotated synonymy dataset described in section. \ref{umls-atom-strings}.

\subsubsection{UBERT-B1 and UBERT-B2}
These two UBERT variants are similar in the following sense. Both models are initialized on random weights and further pretrained on MLM task and then the resulting checkpoint from training the MLM task is used consecutively to initialize the pretraining of SP task with annotated synonymy dataset \ref{umls-atom-strings}.\\
The difference between UBERT-B1 and UBERT-B2 lies in the datasets used for pretraining the MLM task. In UBERT-B1 the MLM task is pretrained using the combined dataset of UMLS atom strings and biomedical literature. 
In UBERT-B2, the MLM task is pretrained using only UMLS atom strings (see section \ref{umls-atom-strings}). 
Once the models are trained on the MLM task with different datasets, each resulting model is then used to initialize the weights for further pretraining with the synonymy prediction task using the annotated synonymy dataset in \ref{synonymy-dataset} (which use the same resources and the input sequence length as the UBERT-A model).

\subsection{\label{other-uberts}Pretrained Variants}
In pretrained variants the UBERT-A model is further pretrained on top of already trained weights of four BERT-based models.\\
We initialize each pretrained variant with the pretrained weights and the tokenizer released by the corresponding BERT-based model, and further pretrain using the annotated synonymy dataset on SP task with the same hardware requirements as UBERT-A and the same maximum input sequence length.\\

All the UBERT variants and pretrained variants are tested on the synonymy prediction task using the test dataset from Table \ref{tab:annotated-synonymy-dataset}. The best performing model (with regard to F1-score) is selected from the training epochs or steps for each experiment. Testing is done on this best performing model.

\subsection{Implementation Details}
We use the Transformers\footnote{\url{https://huggingface.co/transformers/v4.5.1/index.html}} API to develop the training, evaluation and testing scripts of all the models mentioned in this paper. Since training, evaluation and testing of BERT-based architectures with millions of data points, is computationally expensive, we do distributed training, evaluation and testing utilizing the Pytorch\footnote{\url{https://pytorch.org/}} framework. The physical infrastructure used for the experiments is the Biowulf high-performance computing cluster\footnote{\url{https://hpc.nih.gov/}} at the National Institute of Health (NIH). We use Slurm\footnote{\url{https://slurm.schedmd.com/documentation.html}} workload manager to submit the training, evaluation and testing jobs to Biowulf. If not stated specifically, all the training parameters are set to defaults as mentioned in Transformers API documents\footnote{\url{https://huggingface.co/transformers/v4.5.1/main\_classes/trainer.html}} (e.g., learning rate, gradient accumulation steps, optimizer, etc.). Table \ref{tab:computation-resources} summarizes the computing resources required by the models.\\
Our code will be available at \url{https://github.com/naaclubert/UBERT}. We recommend reaching out to Nguyen et al. \cite{10.1145/3442381.3450128} for training and testing data.

\subsection{Semi-quantitative Evaluation}
We divide the large testing dataset into 10 subsets based on the degree of lexical similarity (measured by the Jaccard score based on normalized words) between the pairs of atoms being evaluated for synonymy. Since the Jaccard score varies between 0 and 1, we use 10 intervals of 0.10. Using the best performing UBERT model, we compute the usual performance metrics (precision, recall and F1 score) for the pairs of atoms in each interval of lexical similarity.

\subsection{Statistical Analysis}
To assess the statistical significance of the difference in overall
performance between the best UBERT and the reference LexLM on the GEN\_ALL dataset, we perform a McNemar test. This test compares the distribution of positive and negative predictions between the two models.

\begin{table*}[ht]
\begin{center}
\resizebox{0.8\textwidth}{!}{%
\begin{tabular}{@{}cccccc@{}}
\toprule
Variant Category & \multicolumn{1}{l}{Model} & \multicolumn{1}{l}{Best F1} & \multicolumn{1}{l}{Precision} & \multicolumn{1}{l}{Recall} & \multicolumn{1}{l}{Accuracy} \\ \midrule
\multirow{5}{*}{Baseline} & LexLM & 0.9061 & 0.8875 & 0.9254 & 0.9938 \\ \cmidrule(l){2-6} 
 & SapBERT & 0.0538 & 0.0286 & 0.4484 & 0.4882 \\ \cmidrule(l){2-6} 
 & UmlsBERT & 0.0617 & 0.0325 & 0.6093 & 0.3983 \\ \cmidrule(l){2-6} 
 & BioBERT & 0.0688 & 0.0361 & 0.7421 & 0.3479 \\ \cmidrule(l){2-6} 
 & BlueBERT & 0.0818 & 0.0428 & 0.9479 & 0.3098 \\ \midrule
\multirow{3}{*}{UBERT Variant} & UBERT-A & 0.9319 & 0.8920 & 0.9756 & 0.9954 \\ \cmidrule(l){2-6} 
 & UBERT-B1 & 0.9316 & 0.8935 & 0.9731 & 0.9954 \\ \cmidrule(l){2-6} 
 & UBERT-B2 & 0.9340 & 0.8963 & 0.9749 & 0.9955 \\ \midrule
\multirow{4}{*}{Pretrained Variant} & SapBERT + UBERT & \underline{0.9420} & \underline{0.9089} & \underline{0.9775} & \underline{0.9961} \\ \cmidrule(l){2-6} 
 & UmlsBERT+UBERT & 0.9351 & 0.8977 & 0.9757 & 0.9956 \\ \cmidrule(l){2-6} 
 & BioBERT+UBERT & 0.9376 & 0.9018 & 0.9764 & 0.9958 \\ \cmidrule(l){2-6} 
 & BlueBERT+UBERT & 0.9391 & 0.9041 & 0.9768 & 0.9959 \\ \bottomrule
\end{tabular}%
}
\caption{Results for all the experimented models. Models are categorized into three groups. The baseline category consists of the previous LexLM baseline and BERT-based models tested for the UVA task (without any pretraining or fine-tuning). The UBERT Variant category consists of the three UBERT variants. The pretrained Variant category lists the results for BERT-based models further pretrained using UBERT.}
\label{tab:all-results}
\end{center}
\end{table*}

\section{Results}

\vspace{-1mm}
Table \ref{tab:all-results} consolidates the best F1-score, precision, recall and accuracy values for all the models. We categorize models into three categories, baselines (LexLM and "off-the-shelf" biomedical BERT models), UBERT variants and pretrained variants.\\

\subsection{Overall Performance of UBERT}
As shown in Table \ref{tab:all-results}, SapBERT+UBERT shows a significant performance improvement over the LexLM. The McNemar statistics (5615042.0) indicates that the difference is statistically significant (p < 0.001). SapBERT+UBERT also outperforms all the "off-the-shelf" BERT-based baselines.
\subsection{UBERT Variants}
Among the three UBERT variants, UBERT-B2 perform slightly better than the other two variants in the same category indicating the MLM task pretrained using UMLS data has a positive impact on the training of UBERT.\\
\subsection{Pretrained Variants}
The results in the pretrained variant category in Table \ref{tab:all-results} show that further pretraining of the BERT-based models using UBERT improves the performance of these models on the UVA task.\\

\subsection{Semi-quantitative Evaluation}
\begin{figure}[ht]
     \centering
     \includegraphics[width=8cm]{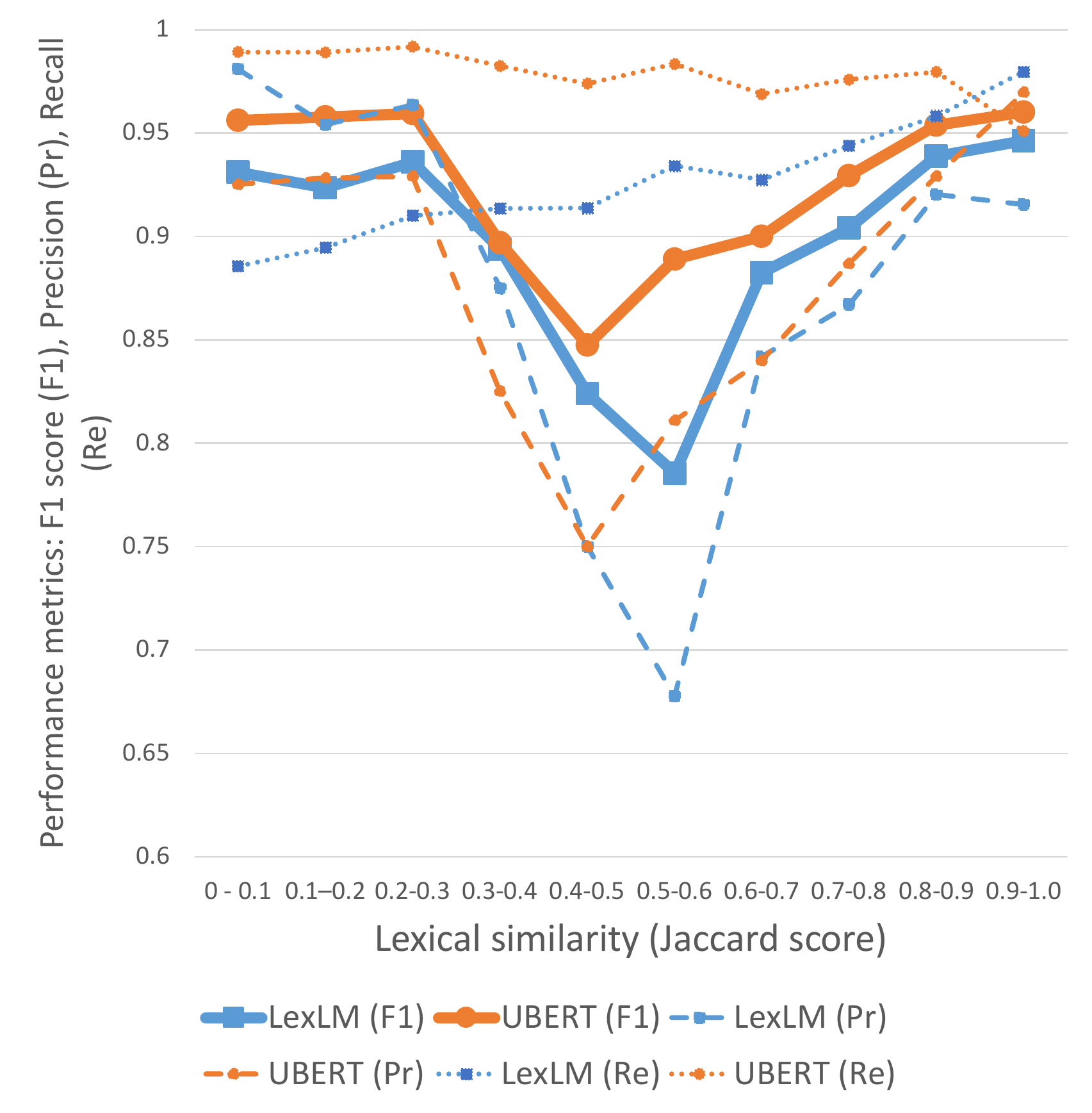}
     \caption{Performance across levels of lexical similarity}
     \label{fig:semi-quantitative}
\end{figure}

As shown in Figure \ref{fig:semi-quantitative}, the F1-score is consistently higher for UBERT compared to the LexLM baseline, at all levels of lexical similarity. The same can be said of recall, with the exception of the highest level of lexical similarity, where LexLM performs better. For precision, however, LexLM performs better at low levels of lexical similarity, whereas UBERT performs better at medium and high levels of lexical similarity.\\

\section{Discussion}

\subsection{Findings and Insights}
\textbf{Overall performance.} We find that UBERT further pretrained on SapBERT significantly outperforms the previous SOTA for the UVA task interms of F1-score (+3.6\%), precision (+2.1\%) and recall (+5.2\%). Such performance gains will decrease the number of false positives and false negatives, reducing the need for manual curation.\\
\textbf{MLM training data.} We find that training on the MLM task improves the performance of  UBERT if only UMLS data is used to train the MLM task. If UMLS data and the biomedical literature are used in combination, the performance drops. As we have indicated in section \ref{tokenizer}, most of the terms present in the biomedical literature are not found in the UMLS Metathesaurus and we believe that those words might act as noise rather than added knowledge, confusing the model. When including knowledge to improve the performance of a language model, it is important to make sure the data align well with the specific task and have enough coverage.\\
\textbf{Knowledge transfer.} We find that BERT-based models, although previously trained on biomedical, clinical and UMLS data, perform poorly on the UVA task if not further pretrained. This poor performance indicates that the knowledge gained by these models is not directly transferable to the UVA task.\\
\textbf{Best performing models.} We also find that among these BERT-based models, SapBERT+UBERT performs best. There could be two reasons for this performance. One is that both SapBERT and UBERT are based on the same BERT architecture, therefore well-aligned with each other for better performance. The second reason could be the similarities of the training knowledge between SapBERT and UBERT, namely the fact that both use atom strings, CUIs and synonymy information.\\
\textbf{Generalizability.} A majority of the atom pairs in the testing dataset exhibit low and very low levels of lexical similarity, which reflects the composition of the UMLS Metathesaurus. Therefore, this analysis indicates that UBERT is likely to perform well across the entire UMLS Metathesaurus.
\subsection{Limitations and Future Work}

\textbf{Contextual information.} In this work, the synonymy between terms in a given atom string pair is identified only by incorporating the lexical cues present in the atom strings as identified by the UBERT model. Yet the human experts, when identifying synonymy, incorporate contextual information about the atom strings. We believe that devising a mechanism to incorporate the terminological context of the atom strings (namely hierarchical relations, source synonymy and semantic categorization) would further improve the performance of UBERT.\\
\textbf{Computational cost.} Even though UBERT provides better performance than the baselines, it is extremely expensive to train. We believe that a knowledge distillation approach \cite{gou2021knowledge} could effectively reduce the cost of training. We have not explored the hyper-parameter optimization in this work due to the time taken to train the models. We plan to incorporate hyper-parameter optimization in combination with knowledge distillation in the future.\\
\textbf{Application.} We also plan to test UBERT on several Medical Entity Linking (MEL) tasks since we believe that UVA has elements similar to EL.\\ 
\textbf{Combining models.} We have not combined UBERT with LexLM since the two architectures are vastly different from each other, but we will test an ensemble approach in the future. 
\section{Conclusion}

This work introduces UBERT, a BERT-based language model, pretrained on UMLS terms via a supervised Synonymy Prediction (SP) task replacing the original Next Sentence Prediction (NSP) task, which provides significant performance improvement over the LexLM introduced by \cite{10.1145/3442381.3450128}. Key to its performance are the synonymy prediction task specifically developed for UBERT, the tight alignment of training data to the UVA task, and the similarity of the models used for pretrained UBERT.

\section{Acknowledgments}
 We would like to thank Nguyen et al. for sharing the dataset used in \cite{10.1145/3442381.3450128}.

\bibliography{anthology,custom}
\bibliographystyle{acl_natbib}

\appendix

\end{document}